\documentclass[10pt,twocolumn]{article}
\pdfoutput=1

\usepackage{iccv}
\usepackage{times}
\usepackage{epsfig}
\usepackage{graphicx}
\usepackage{amsmath}
\usepackage{amssymb}
\usepackage{xcolor}
\usepackage{subcaption}
\usepackage{dblfloatfix}
\usepackage{setspace}
\usepackage{bbm}
\usepackage[affil-it]{authblk}

\usepackage[pagebackref=false,breaklinks=true,colorlinks=true,bookmarks=false]{hyperref}

\iccvfinalcopy 

\def\iccvPaperID{****} 

\usepackage{gensymb}
\usepackage{booktabs}
\usepackage{cite}

\definecolor{skyblue3}{RGB}{ 32,  74, 135}
\definecolor{scarletred3}{RGB}{164,   0,   0}
\definecolor{aluminium3}{RGB}{186, 189, 182}
\definecolor{aluminium5}{RGB}{ 85,  87,  83}
\definecolor{chameleon3}{RGB}{ 78, 154,   6}

\ificcvfinal\fi
\begin{document}
	
\title{Center3D: Center-based Monocular 3D Object Detection with Joint Depth Understanding}

\author{Yunlei Tang\textsuperscript{1}\thanks{harryyunlei@gmail.com}
	\qquad
	Sebastian Dorn
	\qquad
	Chiragkumar Savani}
\affil{\textsuperscript{1}Department of Electrical Engineering and Information Technology,\\ Technische Universit\"at Darmstadt, Germany}

\maketitle

\begin{abstract}
	
Localizing objects in 3D space and understanding their associated 3D properties is challenging given only monocular RGB images. 
The situation is compounded by the loss of depth information during perspective projection.
We present Center3D, a one-stage anchor-free approach, to efficiently estimate 3D location and depth using only monocular RGB images.
By exploiting the difference between 2D and 3D centers, we are able to estimate depth consistently.
Center3D uses a combination of classification and regression to understand the hidden depth information more robustly than each method alone.
Our method employs two joint approaches: (1) \textbf{LID}: a classification-dominated approach with sequential \textbf{L}inear \textbf{I}ncreasing \textbf{D}iscretization.
(2) \textbf{DepJoint}: a regression-dominated approach with multiple Eigen's transformations~\cite{eigen2014depth} for depth estimation.
Evaluating on KITTI dataset~\cite{geiger2012we} for moderate objects, Center3D improved the AP in BEV from $29.7\%$ to $\mathbf{42.8\%}$, and the AP in 3D from $18.6\%$ to $\mathbf{39.1\%}$.
Compared with state-of-the-art detectors, Center3D has achieved the best speed-accuracy trade-off in realtime monocular object detection.

\end{abstract}

\begin{figure*}[t]
	\centering
	\includegraphics[width=0.95\linewidth]{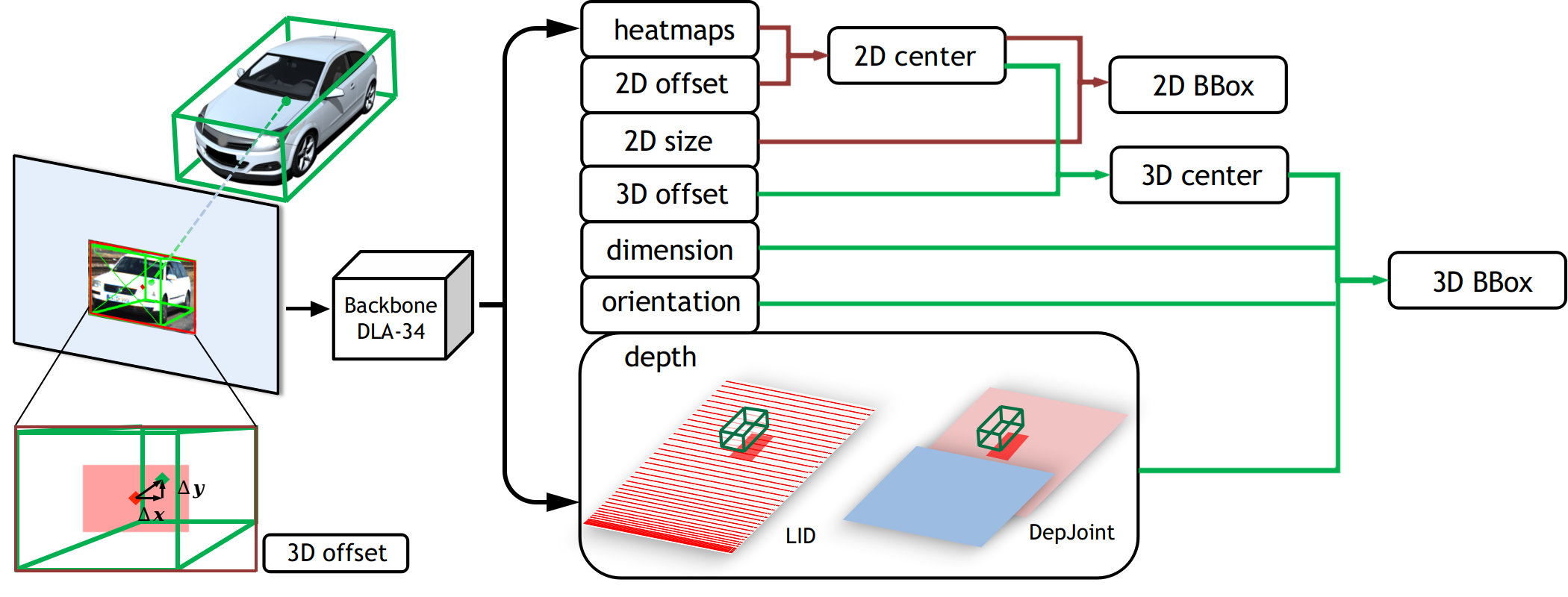}
	\caption{Overview of Center3D. A monocular input image is fed to the backbone {\em DLA-34}, which generates feature maps. {\em Heatmaps} and {\em 2D offset} are subsequently used to detect the {\em 2D center}. The latter is relocated by {\em 3D offset} to propose the {\em 3D center}, which is illustrated in the bottom-left of the figure. By applying a combination of regression and classification, {\em DepJoint} or {\em LID}, Center3D is inferring the {\em depth} of the associated {\em 3D center}. {\em Depth}, together with regressed {\em dimensions}, {\em orientation}, and {\em 3D center} are finally used to propose the {\em 3D BBox}.}
	\label{fig:teaser}
\end{figure*}

\section{Introduction}
3D object detection is currently one of the most challenging topics for both industry and academia.
Applications of related developments can easily be found in the areas of robotics, AI-based medical surgery, autonomous driving~\cite{chen2017multi},\cite{rey2002automatic},\cite{surmann2003autonomous} etc.
The goal is to have agents with the ability to identify, localize, react, and interact with objects in their surroundings. 
2D object detection approaches~\cite{law2018cornernet},\cite{liu2016ssd},
\cite{redmon2016you},\cite{ren2015faster},\cite{zhou2019objects} achieved impressive results in the last decade.
In contrast, inferring associated 3D properties from a 2D image turned out to be a challenging problem in computer vision, due to the intrinsic scale ambiguity of 2D objects and the lack of depth information. 
Hence many approaches involve additional sensors like LiDAR~\cite{shi20193d},\cite{shin2019roarnet},\cite{wang2019frustum} or radar~\cite{moosmann2009segmentation},\cite{vasile2005pose} to measure depth. 
Modern LiDAR sensors have proven themselves a reliable choice with high depth accuracy.
On the other hand, there are reasons to prefer monocular-based approaches too.
LiDAR has reduced range in adverse weather conditions, while visual information of a simple RGB camera is more dense and also more robust under rain, snow, etc. 
Another reason is that cameras are currently significantly more economical than high precision LiDARs and are already available in {\em e.g.} robots, vehicles, etc.
Additionally the processing of single RGB images is much more efficient and faster than processing 3D point clouds in terms of CPU and memory utilization.

These compelling reasons have led to research exploring the possibility of 3D detection solely from monocular images~\cite{brazil2019m3d},\cite{chen2016monocular},\cite{jorgensen2019monocular},\cite{li2019gs3d},\cite{mousavian20173d},\cite{qin2019monogrnet},\cite{roddick2018orthographic},\cite{simonelli2019disentangling}.
The network structure of most 3D detectors starts with a 2D region proposal based (also called anchor based) approach, which enumerates an exhaustive set of predefined proposals over the image plane and classifies/regresses only within the region of interest (ROI).
Mousavian {\em et al.}~\cite{mousavian20173d} uses a two-stage 2D detector and adds specific heads for regression of 3D properties. 
The resulting 3D cuboid is then fine tuned to ensure that it tightly fits inside the associated 2D bounding box.
GS3D~\cite{li2019gs3d} modifies Faster RCNN to propose a 3D guidance based on a 2D bounding box and orientation, which guides to frame a 3D cuboid by refinement.
OFT-Net~\cite{roddick2018orthographic} maps 2D image features into an orthographic bird's-eye view (BEV) by an orthographic feature transformation, and infers the depth in a reprojected 3D space.
Mono3D~\cite{chen2016monocular} focuses on the generation of proposed 3D boxes, which are scored by features like contour and shape under the assumption that all vehicles are placed on the ground plane.
MonoGRNet~\cite{qin2019monogrnet} consists of parameter-specific subnetworks.
All further regressions are guided by the detected 2D bounding box.
M3D-RPN~\cite{brazil2019m3d} demonstrates a single-shot model with a standalone 3D RPN, which generates 2D and 3D proposals simultaneously. 
Additionally, the specific design of depth-aware convolutional layers improved the network's 3D understanding.
With the help of an external network, Multi-Fusion~\cite{xu2018multi} estimates a disparity map and subsequently a LiDAR point cloud to improve 3D detection.
Due to multi-stage or anchor-based pipelines, most of them perform slowly.

Most recently, to overcome the disadvantages above, 2D anchor-free approaches have been used by researchers~\cite{chen2016monocular},\cite{duan2019centernet},\cite{law2018cornernet},\cite{zhou2019objects}.
They model objects with keypoints like centers, corners or points of interest of 2D bounding boxes. 
Anchor-free approaches are usually one-stage, thus eliminating the complexity of designing a set of anchor boxes and fine tuning hyperparameters.
The recent work of CenterNet: Objects as Points~\cite{zhou2019objects} proposed a possibility to associate a 2D anchor free approach with a 3D detection.

Nevertheless, the performance of CenterNet is still restricted by the fact that a 2D bounding box and a 3D cuboid are sharing the same center point. 
In this paper we analyze the difference between the center points of 2D bounding boxes and the projected 3D center points of objects, which are almost never at the same image position.
We directly regress the 3D centers from 2D centers to locate the objects in the image plane and in 3D space separately.
Furthermore, we show that the weakness of monocular 3D detection is mainly caused by imprecise depth estimation.
This causes the performance gap between LiDAR-based and monocular image-based approaches.
By examining depth estimation in monocular images, we show that a combination of classification and regression explores visual clues better than using only a single approach. 
An overview of our approach is shown in Figure~\ref{fig:teaser}.

We introduce two approaches to validate this conclusion: (1) Motivated by DORN~\cite{fu2018deep} we consider depth estimation as a sequential classification with residual regression. 
According to the statistics of the  instances in the KITTI dataset, a novel discretization strategy is used.
(2) We divide the whole depth range of objects into two bins, foreground and background,  either with overlap or associated. 
Classifiers indicate which depth bin or bins the object belongs to. 
With the help of Eigen's transformation~\cite{eigen2014depth}, two regressors are trained to gather specific features for closer and farther away objects, respectively.
For illustration see the depth part in Figure~\ref{fig:teaser}.

Compared to CenterNet, our approach improved the AP of easy, moderate, hard objects in BEV from $31.5$, $29.7$, $28.1$ to $\mathbf{55.8}$, $\mathbf{42.8}$, $\mathbf{36.6}$, in 3D space from $19.5$, $18.6$, $16.6$ to $\mathbf{49.1}$, $\mathbf{38.9}$, $\mathbf{33.5}$, which is comparable with state-of-the-art approaches. 
Center3D achieves the best speed-accuracy trade-off on the KITTI dataset in the field of monocular 3D object detection. 
Details are given in Table~\ref{tab:sota}.

\begin{table*}[t]
	\centering
	\begin{tabular}{l c c c c c c c}
		\hline
		& & \multicolumn{3}{c}{2D AP} & \multicolumn{3}{c}{BEV / 3D AP}  \\
		& RT & Easy & Mode & Hard & Easy & Mode & Hard \\
		\hline
		CenterNet~\cite{zhou2019objects} & \underline{50} & \underline{97.1} & 87.9 & \textbf{79.3} & 31.5 / 19.5 & 29.7 / 18.6 & 28.1 / 16.6 \\
		CenterNet (ct3d) & & 87.1 & 85.6 & 69.8 & 46.8 / 39.9 & 37.9 / 31.4 & 32.7 / 30.1  \\
		\hline
		Mono3D~\cite{chen2016monocular} & - & 92.3 & \textbf{88.7} & 79.0 & 30.5 / 25.2 & 22.4 / 18.2 & 19.2 / 15.5  \\
		MonoGRNet~\cite{qin2019monogrnet} & 60 & - & - & - & - / \textbf{50.5} & - / 37.0 & - / 30.8  \\
		Multi-Fusion~\cite{xu2018multi} & 120 & - & - & - & 55.0 / 47.9 &  36.7 / 29.5 & 31.3 / 26.4  \\
		M3D-RPN~\cite{brazil2019m3d} & 161 & 90.2 & 83.7 & 67.7 & \underline{55.4} / 49.0 & \underline{42.5} / \textbf{39.6} & 35.3 / 33.0 \\								
		\hline 
		Center3D & \textbf{47} & \textbf{97.2} & \underline{88.0} & \underline{79.2} & 47.7 / 39.9 & 37.8 / 34.5 & 33.0 / 30.6 \\
		Center3D (+ra) & 52 & 96.4 & 87.0 & 78.7 & 54.3 / \underline{49.8} & 42.1 / \underline{39.1} & \textbf{40.2} / \textbf{34.0} \\
		Center3D (+lid) & 53 & 96.9 & 87.5 & 79.0 & 51.3 / 44.0 & 39.3 / 35.0 & 33.9 / 30.6 \\
		Center3D (+dj+ra) & 54 & 95.0 & 78.8 & 70.0 & \textbf{55.8} / 49.1 & \textbf{42.8} / 38.9 & \underline{36.6} / \underline{33.6} \\
		\hline
	\end{tabular}
	\caption{Comparison of Center3D with state-of-the-art approaches in 2D and 3D. RT indicates runtime in $ms$. $ct3d$ denotes CenterNet with 3D center points instead of 2D center points. $ra$ indicates a reference area supporting depth and offset estimation, while $dj$ represents the DepJoint approach. The best result is marked in bold, the second best is underlined. See Sec.~\ref{sec:sota} for a detailed discussion. AP values are given in percentage.}
	\label{tab:sota}
\end{table*}

\section{Center3D}

\subsection{Baseline and Motivation}
\label{sec:baseline}
\begin{figure*}[t]
	\centering
	\begin{tabular}{l}
		\includegraphics[width=0.5\textwidth]{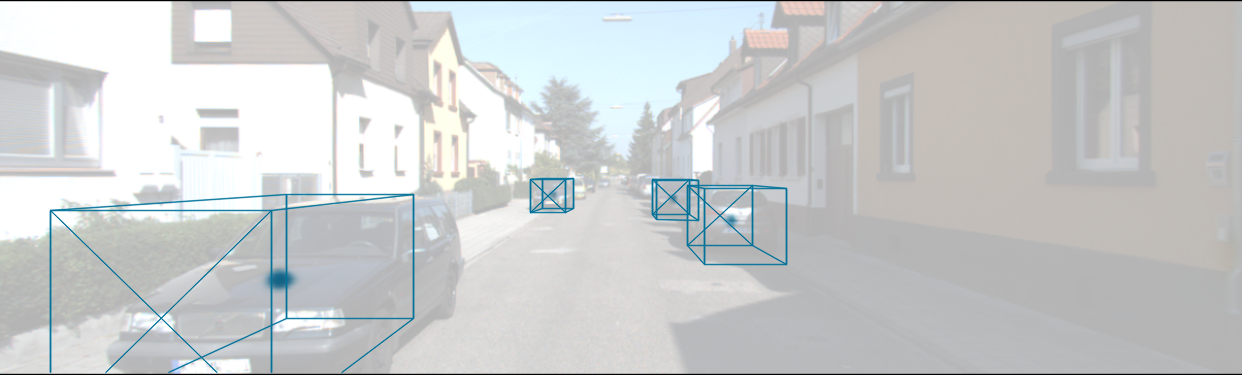}
		\includegraphics[width=0.5\textwidth]{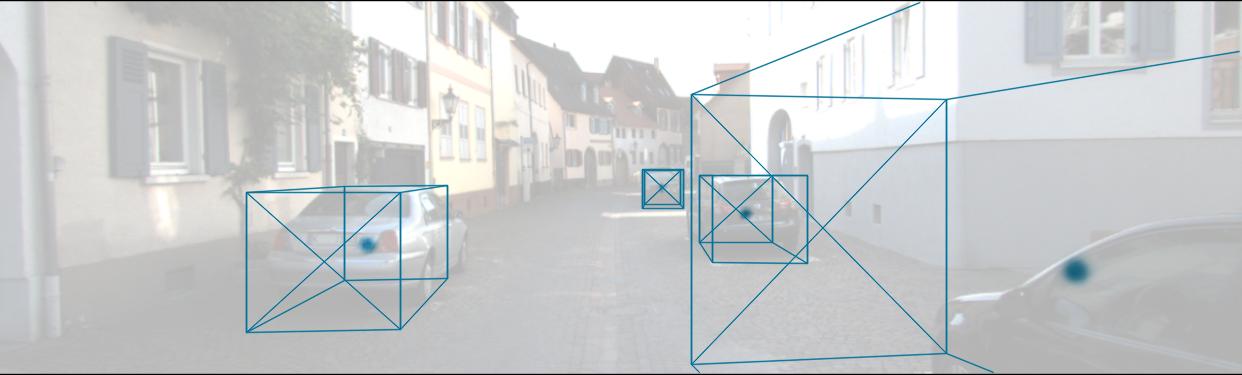}	\\
		\includegraphics[width=0.5\textwidth]{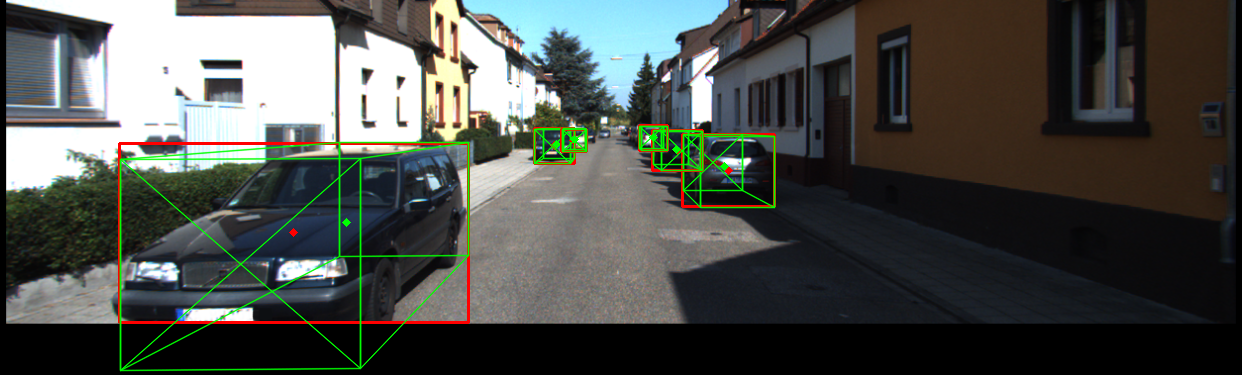}
		\includegraphics[width=0.5\textwidth]{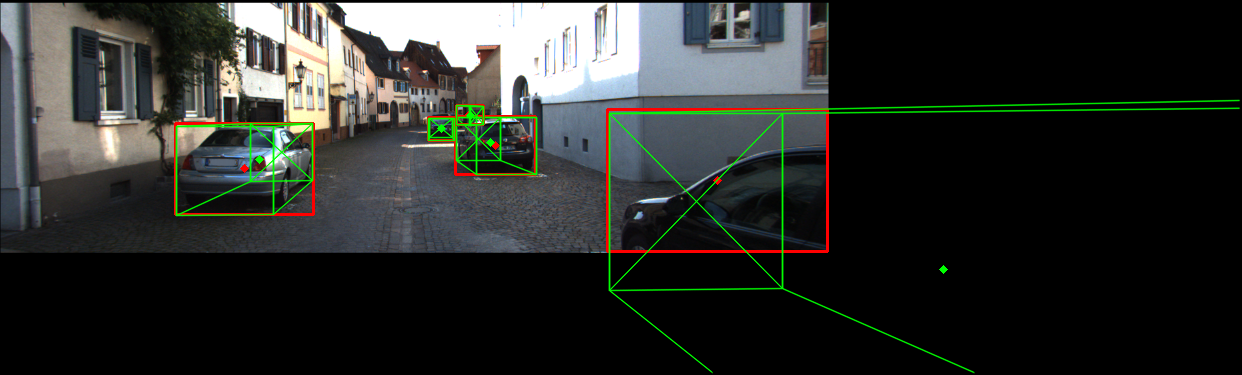} 	\\
		\includegraphics[width=0.5\textwidth]{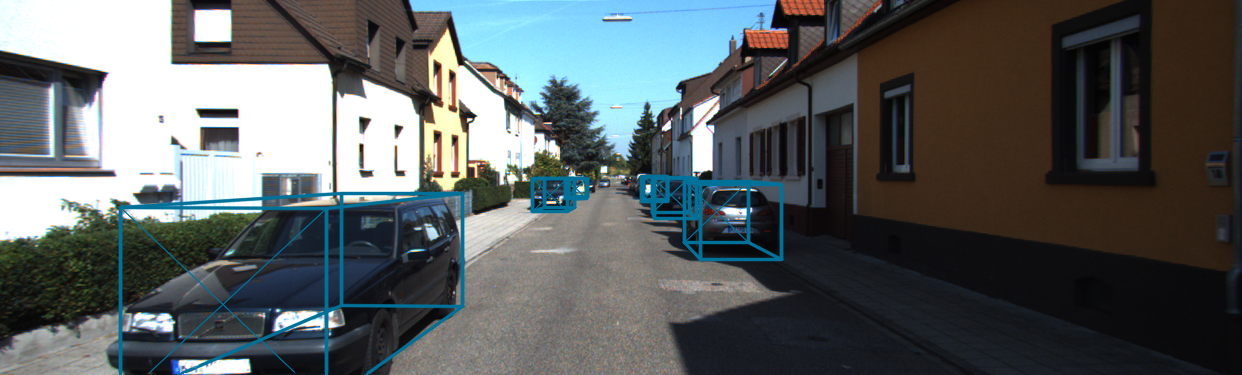}
		\includegraphics[width=0.5\textwidth]{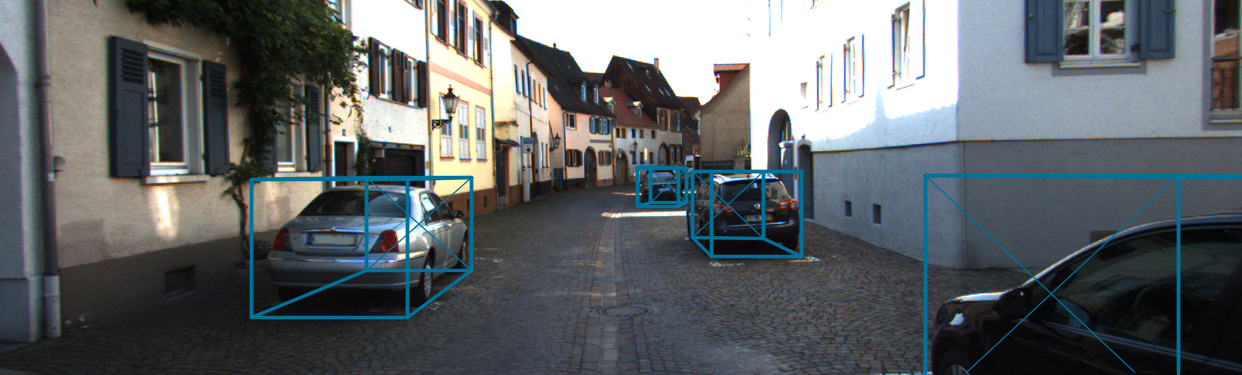} 	\\
	\end{tabular}
	\caption{3D bounding box estimation on KITTI validation set. \emph{first row:} projected 3D bounding boxes located around estimated center points of 2D bounding boxes. The position of centers is generated by the peak of the Gaussian kernel on the heatmap. \emph{second row:} the ground truth of input images. Here the 2D (red) and 3D (green) projected bounding boxes with their center points are shown. To visualize the annotations as complete as possible the images exhibit black paddings (application of affine transformations). As shown, a proper estimation of the difference between the center points of 2D bounding boxes and objects in 3D space is crucial, especially when vehicles are heading to the image boundary. \emph{third row:} the output of Center3D. The 3D cuboid is based on 3D center points shifted from 2D space with offset. }
	\label{fig:baseline}
\end{figure*}

\begin{figure*}[t]
	\centering
	\begin{subfigure}[t]{\linewidth}
		\centering
		\includegraphics[width=0.22\linewidth]{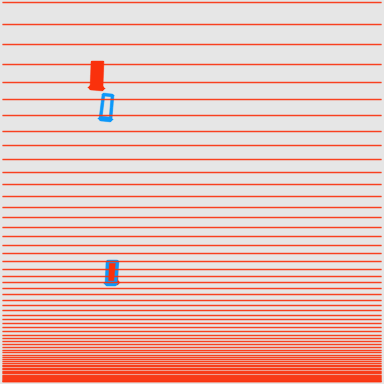}
		\includegraphics[width=0.22\linewidth]{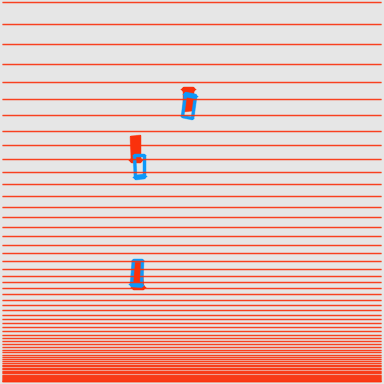}
		\includegraphics[width=0.22\linewidth]{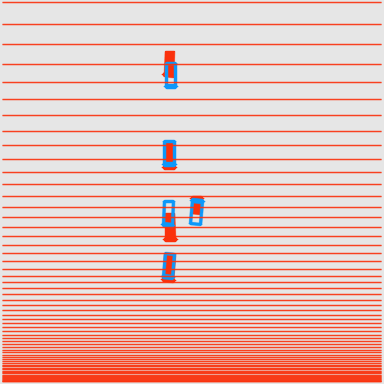}
		\includegraphics[width=0.22\linewidth]{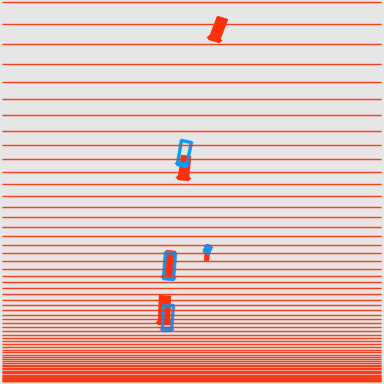}
		\caption*{SID from BEV}
	\end{subfigure}\\
	\begin{subfigure}[t]{\linewidth}
		\centering
		\includegraphics[width=0.22\linewidth]{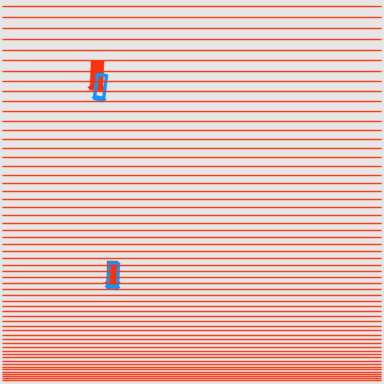}
		\includegraphics[width=0.22\linewidth]{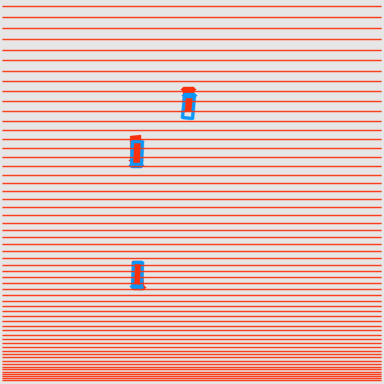}
		\includegraphics[width=0.22\linewidth]{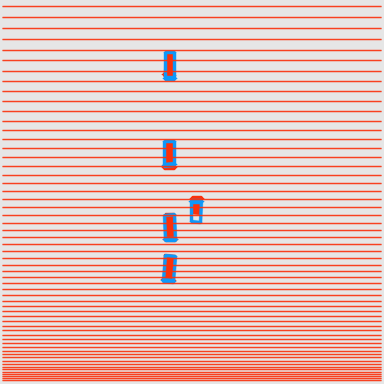}
		\includegraphics[width=0.22\linewidth]{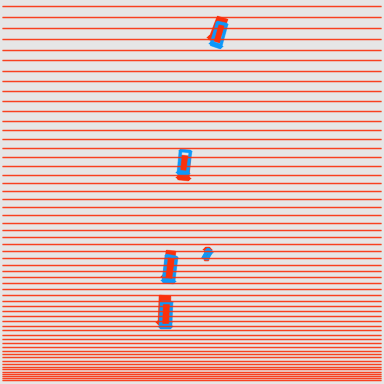}
		\caption*{LID from BEV}
	\end{subfigure}
	\caption{The comparison of the discretization strategies LID ({\em first row}) and SID ({\em second row}) from BEV between $0 m$ and $54 m$, with a setting of $d_\text{min} = 1 m$, $d_\text{max} = 91 m$ and $N = 80$. The solid red lines indicate the threshold of each bin, the solid rectangles represent the ground truth vehicles in BEV, while blue rectangles represent the estimations.}
	\label{fig:lid}
\end{figure*}

The 3D detection approach of CenterNet described in~\cite{zhou2019objects} is the basis of our work. 
It models an object as a single point: the center of its 2D bounding box.
For each input monocular RGB image, the original network produces a heatmap for each category, which is trained with focal loss~\cite{lin2017focal}. 
The heatmap describes a confidence score for each location, the peaks in this heatmap thus represent the possible keypoints of objects. 
All other properties are then regressed and captured directly at the center locations on the feature maps respectively. 
For generating a complete 2D bounding box, in addition to width and height, a local offset will be regressed to capture the quantization error of the center point caused by the output stride. 
For 3D detection and localization, the additional abstract parameters, i.e. depth, 3D dimensions and orientation, will be estimated separately by adding a head for each of them.
However, the reconstruction of the lacking space information is ill-posed and challenging because of the inherent ambiguity~\cite{fu2018deep},\cite{geiger2012we}. 
Following the output transformation of Eigen et al.~\cite{eigen2014depth} for depth estimation, CenterNet converts the feature output into an exponential area to suppress the depth space. 

In Table~\ref{tab:sota}, we show the reproduced evaluation results of CenterNet on the KITTI dataset, which splits all annotated objects into easy, moderate and hard targets.
The results are extended by the approaches introduced in Section~\ref{sec:depth} and \ref{sec:offset3d}.
As CenterNet, we also only focus on the performance in vehicle detection according to standard training and validation splits in literature~\cite{chen20153d}. 
To numerically compare our results with other approaches we use intersection over union (IoU) based on 2D bounding boxes (AP), orientation (AOP), and bounding boxes in Bird's-eye view (BEV AP). 
From the first row in Table~\ref{tab:sota}, we see that the 2D performance of CenterNet is very good.
In contrast, the APs in BEV and especially in 3D perform poorly. 
This is caused by the difference between the center point of the visible 2D bounding box in the image and the projected center point of the complete object from physical 3D space. 
This is illustrated in the first two rows of Figure~\ref{fig:baseline}. 
Here a qualitative comparison and demonstration of the influences, caused by the shift of center locations during different driving scenarios, is shown. 
A center point of the 2D bounding box for training and inference is enough for detecting and decoding 2D properties, {\em e.g.} width and height, while all additionally regressed 3D properties, {\em e.g.} depth, dimension and orientation,  should be consistently decoded from the projected 3D center of the object. 
The gap between 2D and 3D decreases for faraway objects and for objects which appear in the center area of the image plane.
However the gap between 2D and 3D center points becomes significant for objects that are close to the camera or on the image boundary. 
Due to perspective projection, this offset will increase as vehicles get closer. 
Close objects are especially important for technical functions based on perception (e.g. in autonomous driving or robotics).

\subsection{Enriching Depth Information}
\label{sec:depth}
This section first introduces two novel approaches to infer depth cues over monocular images: First, we adapt the advanced DORN~\cite{fu2018deep} approach from pixel-wise to instance-wise depth estimation.
We introduce a novel linear-increasing discretization (\textbf{LID}) strategy to divide continuous depth values into discrete ones, which distributes the bin sizes more evenly than spacing-increasing discretization (SID) in DORN.
Additionally, we employ a residual regression for refinement of both discretization strategies.
Second, with the help of a reference area (\textbf{RA}) we describe the depth estimation as a joint task of classification and regression (\textbf{DepJoint}) in exponential range.

\subsubsection{LID}

\begin{figure}
	\centering
	\begin{tabular}{c}
		\includegraphics[width=0.9\linewidth]{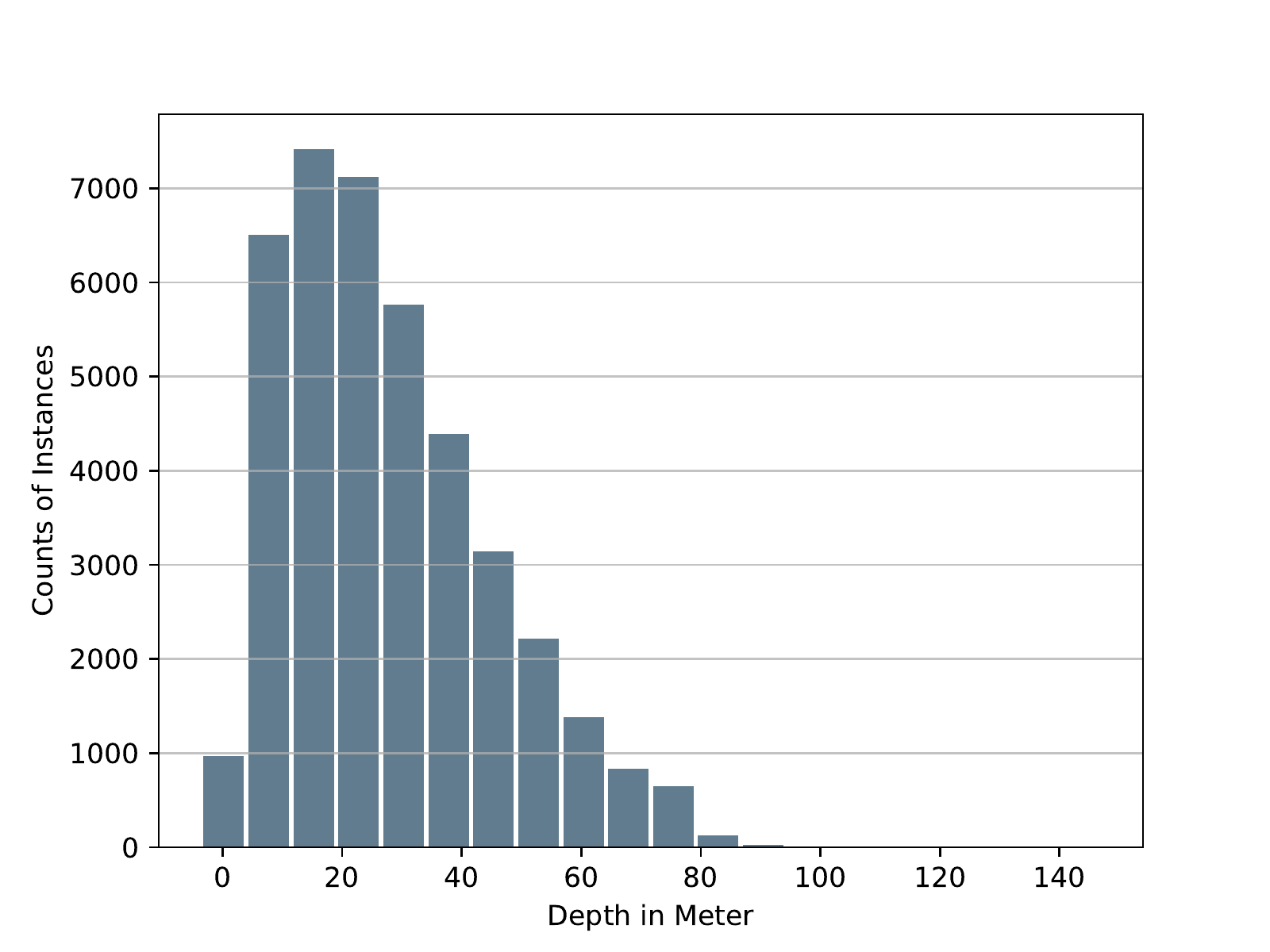}
	\end{tabular}
	\vspace{-2.5mm}
	\caption{Histogram of the depth. The analysis is based on instances in the KITTI dataset.}
	\label{fig:kitti_hist}
\end{figure}

Usually a faraway object with higher depth value and less visible features will induce a higher loss, which could dominate the training and increases uncertainty. 
On the other hand these targets are usually less important for functions based on object detection.
This is also the motivation for the SID strategy to discretize the given continuous depth interval $[d_\text{min}, d_\text{max}]$ in {\em log} space and hence down-weight farther away objects, see Eq.~\ref{eq:lid}.
However, such a discretization often yields too dense bins within unnecessarily close range, where objects barely appear (as shown in Figure~\ref{fig:lid} first row). 
According to the histogram in Figure~\ref{fig:kitti_hist} most instances of the KITTI dataset are between $5\ m$ and $80\ m$. 
Assuming that we discretize the range between $d_\text{min} = 1\ m$ and $d_\text{max} = 91\ m$ into $N = 80$ sub-intervals, $29$ bins will be involved within just $5\ m$. 
Thus, we use the LID strategy to ensure the lengths of neighboring bins increase linearly instead of {\em log}-wise.
For this purpose, assume the length of the first bin is $\delta$.
Then the length of the next bin is always $\delta$ longer than the previous bin.   
Now we can encode an instance depth $d$ in $l_\text{int} = \lfloor l \rfloor$ ordinal bins according to LID and SID respectively.
Additionally, we reserve and regress the residual decimal part $l_\text{res} = l - l_\text{int}$ for both discretization strategies:

\begin{equation}
\begin{split}
\text{SID: \quad}    l  & = N \frac{\log d - \log d_\text{max}}{\log d_\text{max} - \log d_\text{min}},
\\
\text{LID: \quad}	 \delta & = \frac{2 (d_\text{max} - d_\text{min})}{N(1 + N)}, \\
l	& = - 0.5 + 0.5 \sqrt{1 + \frac{8(d - d_\text{min})}{\delta}}.
\end{split}
\label{eq:lid}
\end{equation}

During the inference phase, DORN counts the number of activated bins with probability higher than 0.5, as estimated by the ordinal label $\hat{l}_\text{int}$, and uses the median value of the $\hat{l}_\text{int}$-th bin as the estimated depth in meters. 
The notation of symbols with $\hat{\cdot}$ denotes the output of estimation.
However, relying on discrete median values of bins only is not precise enough for instance localization. 
Hence we modify the training to be a combination of classification and regression. 
For classification we follow the ordinal loss with binary classification and add a shared layer to regress the residuals $l_\text{res}$ additionally.
Given an input RGB image $I \in \mathbb{R}^{W \times H \times 3}$, where $W$ represents the width and $H$ the height of $I$, we generate a depth feature map $\hat D \in \mathbb{R}^{\frac{W}{R} \times \frac{H}{R} \times (2N + 1)}$, where $R$ is the output stride. Backpropagation is only applied on the centers of 2D bounding boxes located at $\mathbf{\hat c_\text{2D}^i} = (\hat x_\text{2D}^i, \hat y_\text{2D}^i)$, where $i \in \{0, 1, ..., K-1\}$ indicates the instance number of total K instances over the image.
The final loss $\mathcal{L}_\text{dep}$ is defined as the sum of the ordinal loss $\mathcal{L}_\text{ord}^{i}$ and residuals loss $\mathcal{L}_\text{res}^{i}$:
\begin{equation}
\begin{split}
\mathcal{L}_\text{dep} &= \sum_{i = 0}^{K - 1} (\mathcal{L}_\text{ord}^{i} + \mathcal{L}_\text{res}^{i}),
\\
\mathcal{L}_\text{ord}^{i} &= - \left(\sum_{n=0}^{l^i - 1} \log \mathcal{P}_n^i + \sum_{n=l^i}^{N - 1}\log(1 - \mathcal{P}_n^i) \right),
\\
\mathcal{P}_n^i &= \mathcal{P}\left(\hat{l}^{i} > n\right),
\\
\mathcal{L}_\text{res}^{i} &= \text{SmL1}(\hat{l}_\text{res}^{i}, l_\text{res}^{i}),
\end{split}
\label{eq:lidloss}
\end{equation}
where $\mathcal{P}_n^i$ is the probability that the $i$-th instance is farther away than the $n$-th bin, and $SmL1$ represents the smooth L1 loss function~\cite{paszke2017automatic}.
During inference, the amount of activated bins will be counted up as $\hat l_\text{int}^i$. 
We refine the result by taking into account the residual part, $\hat l = \hat l_\text{int}^i + \hat l_\text{res}^i$, and decode the result by inverse-transformation of Equation~\ref{eq:lid}.

\subsubsection{DepJoint}
\label{sec:depjoint}
\begin{figure}
	\centering
	\begin{tabular}{c}
		\includegraphics[width=0.9\linewidth]{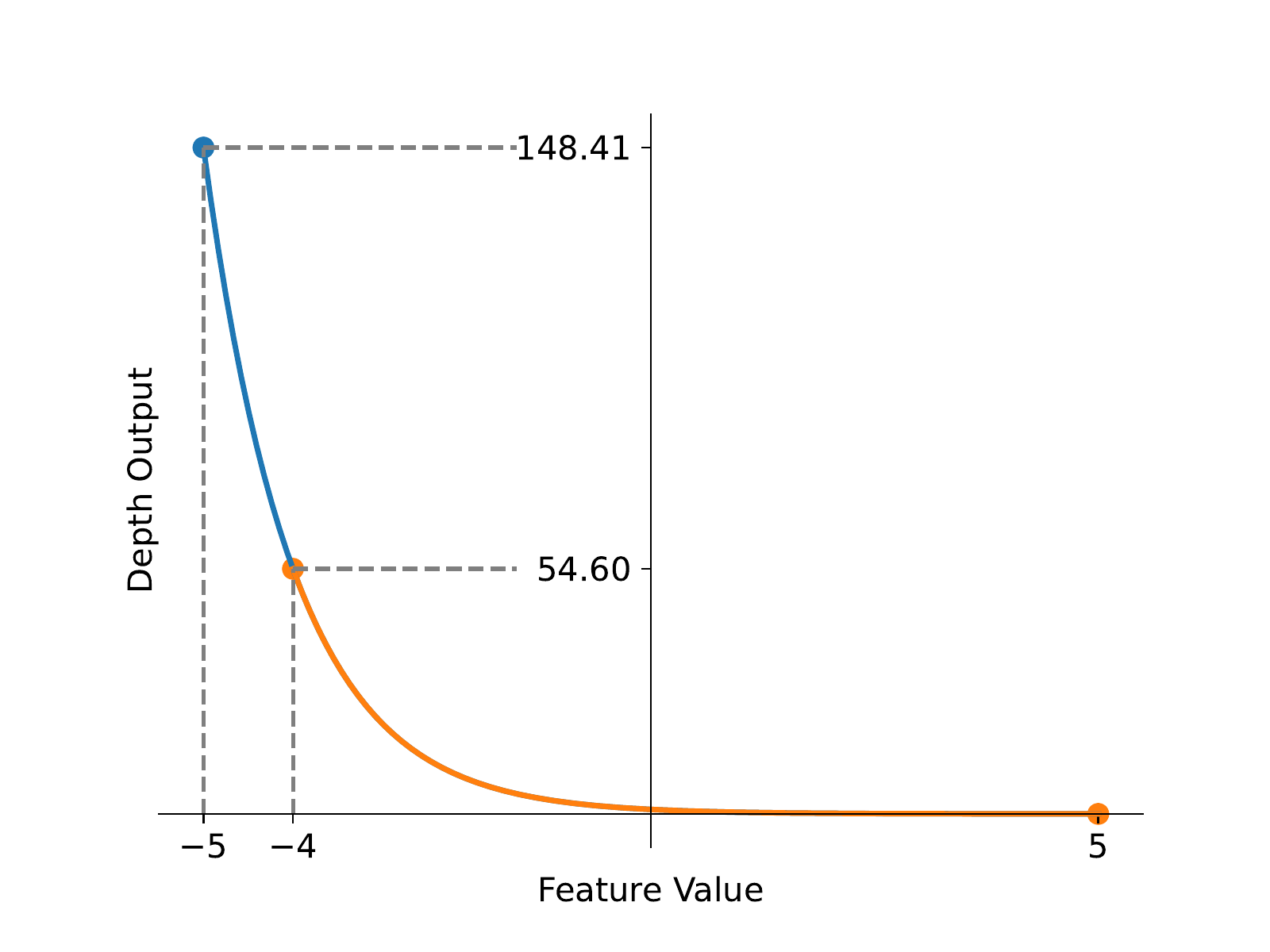}
	\end{tabular}
	\vspace{-2.5mm}
	\caption{Transformation of Eigen et al.~\cite{eigen2014depth} according to depth estimation. The $x$-axis indicates the feature output, and the $y$-axis is the depth output after transformation (given in meter).}
	\label{fig:kitti_eigen}
\end{figure}

The transformation described by Eigen et al.~\cite{eigen2014depth} converts the output depth to an exponential scale. 
It generates a depth feature map $\hat D \in \mathbb{R}^{\frac{W}{R} \times \frac{H}{R} \times 1}$.
The output $\hat d$ at the estimated center point of a 2D bounding box $\mathbf{\hat c_\text{2D}^i} = (\hat x_\text{2D}^i, \hat y_\text{2D}^i)$ is converted to $\Phi(\hat d) = e^{-\hat d}$. 
This enriches the depth information for closer objects by putting more feature values into smaller ranges. 
As shown on the right panel of Figure~\ref{fig:kitti_eigen}, the feature map values between $-4$ and $5$ correspond to a depth up to $54.60 m$, while feature values corresponding to more distant objects up to $148.41 m$ account for only $10\%$ of the feature output range $\left[-5, 5\right]$. 
The transformation is reasonable, since closer objects are of higher importance. 
Eigen's transformation shows an impressive precision on closer objects but disappoints on objects which are farther away.
To improve on the latter, we introduce the DepJoint approach, which treats the depth estimation as a joint classification and regression.
Compared to using Eigen's transformation solely, it emphasizes the distant field.
DepJoint divides the depth range $[d_\text{min}, d_\text{max}]$ in two bins with scale parameter $\alpha$ and $\beta$: 
\begin{equation}
\begin{split}
\text{Bin 1}  &= [d_\text{min}, \ (1 - \alpha) d_\text{min} + \alpha  d_\text{max}],
\\
\text{Bin 2}	 &= [(1 - \beta) d_\text{min} + \beta  d_\text{max}, \ d_\text{max}]. \\
\end{split}
\label{eq:depbins}
\end{equation}
Each bin will only be activated during training when the object lies within the appropriate interval. 
The first bin is used to regress the absolute value of depth $d^i$, while the second bin is used to regress the residual value of depth $\tilde d^i = d_\text{max} - d^i$.
With this transformation, a larger depth value will be supported with more features. 
We use the binary Cross-Entropy loss $CE_b(\cdot)$ for classification of each bin $b$ and regress $d^i$ and $\tilde d^i$ with L1 loss $L1(\cdot)$ subsequent to an output transformation $\Phi(\cdot)$.
Hence the output of the depth head is $\hat D \in \mathbb{R}^{\frac{W}{R} \times \frac{H}{R} \times 6}$ and the loss for training is defined as:
\begin{equation}
\begin{split}
\mathcal{L}_\text{dep} &= \sum_{i = 0}^{K - 1} \left(\mathcal{L}_\text{cls}^{i} + \mathcal{L}_\text{reg}^{i} \right),
\\
\mathcal{L}_\text{cls}^{i} &= \sum_{b} \text{CE}_b (d^i),
\\
\mathcal{L}_\text{reg}^{i} &=  \mathbbm{1}_1 \left(d^i\right) \cdot \text{L1}\left(d^i, \Phi\left(\hat d_1^i\right)\right) 
\\
& \qquad + \mathbbm{1}_2 \left(d^i\right) \cdot \text{L1}\left(d_\text{max} - d^i, \Phi\left(\hat d_2^i\right)\right),
\\
\mathbbm{1}_b \left(d^i\right) &=
\begin{cases}
1, & \text{if } d^i \in \text{Bin }b,\\
0, & \text{else,}
\end{cases}
\end{split}
\label{eq:deploss}
\end{equation}
where $\hat d_b^i$ represents the regression output for the $b$-th bin and $i$-th instance. 
Training is only applied on 2D centers of bounding boxes. 
During inference the weighted average will be decoded as the final result:
\begin{equation}
\begin{split}
\hat d^i &= \mathcal{P}_\text{Bin 1}^i\left(\hat{d}^i\right) \cdot \Phi \left(\hat{d}_1^i\right)
\\
& \qquad + \mathcal{P}_\text{Bin 2}^i\left(\hat{d}^i\right) \cdot \left(d_\text{max} - \Phi \left(\hat{d}_2^i\right)\right),
\\
\end{split}
\label{eq:depdecode}
\end{equation}
where $P_\text{Bin b}^i$ denotes the normalized probability of $\hat{d}^i$.

\subsection{Offset3D: Bridging 2D to 3D}
\label{sec:offset3d}
As described in Section~\ref{sec:baseline}, the significant difference in performance between 2D and 3D object localization results from the gap between the centers of 2D bounding boxes $\mathbf{c_\text{2D}^i} = (x_\text{2D}^i, y_\text{2D}^i)$ and the 3D projected center points of cuboids from physical space $\mathbf{c_\text{3D}^i} = (x_\text{3D}^i, y_\text{3D}^i)$. 
Instinctively we would like to anchor the objects by projected 3D center points $\mathbf{c_\text{3D}^i}$ instead of $\mathbf{c_\text{2D}^i}$. 
Rather than using a regression of width and height $(x_2^i - x_1^i, y_2^i - y_1^i)$ for the 2D task, we can regress the distances from four boundaries to the 3D centers as $(x_\text{3D}^i - x_1^i, x_2^i - x_\text{3D}^i,  y_\text{3D}^i - y_1^i, y_2^i - y_\text{3D}^i)$ to decode the additional 3D parameters properly. 
This strategy has some natural limits, though.
First, the detector can not locate objects very close to the camera with 3D center points outside the image ({\em e.g.} second column in Figure~\ref{fig:baseline}). 
Secondly, there is an ambiguity in the distance with respect to width and height.
Hence we split the 2D and 3D tasks into separate parts. 
We still locate an object with 2D center $\mathbf{c_\text{2D}^i} = (x_\text{2D}^i, y_\text{2D}^i)$, which is definitively included in the image, and determine the 2D bounding box of the visible part with $w^i$ and $h^i$. 
For the 3D task we relocate the projected 3D center $\mathbf{c_\text{3D}^i} = (x_\text{3D}^i, y_\text{3D}^i)$ by adding two head layers on top of the backbone and regress the offset $\mathbf{\Delta c^i} = (x_\text{3D}^i  - x_\text{2D}^i, y_\text{3D}^i - y_\text{2D}^i)$ from 2D to 3D centers.
Given the projection matrix $\mathbf{P}$ in KITTI, we can now determine the 3D location $\mathbf{C} = (X, Y, Z)$ by converting the transformation in homogeneous coordinates.
Similarly, we generate 8 corners of a cuboid by decoding object dimensions based on the 3D location, which is actually the center of the object in the word coordinate system.

\subsection{Reference Area} 
\begin{figure*}
	\centering
	\begin{tabular}{c}
		\includegraphics[width=0.9\linewidth]{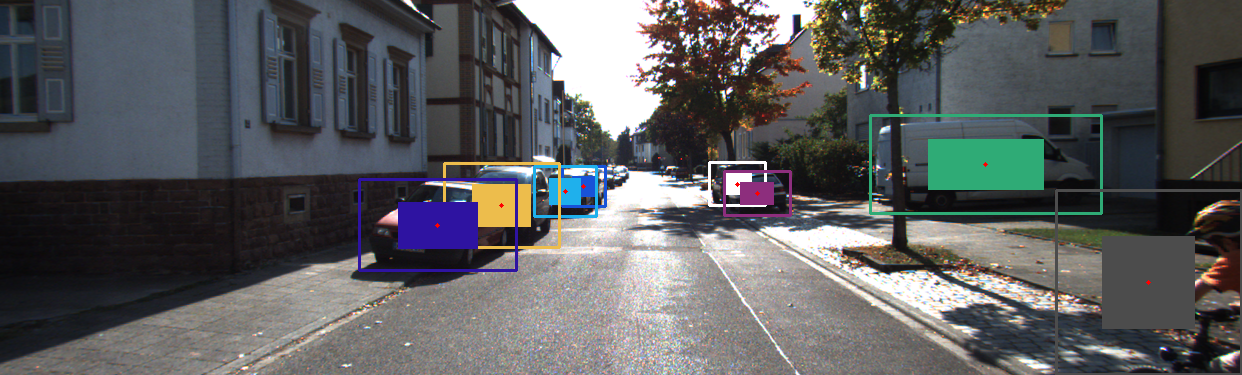}
	\end{tabular}
	\vspace{-2.5mm}
	\caption{Reference area. Instead of relying on single center points of 2D bounding boxes, RAs define scaled rectangles which are centered in 2D bounding boxes and support the inference. The scale $\gamma$ is a hyperparameter and should be chosen properly.}
	\label{fig:ra}
\end{figure*}

Conventionally the regressed values of a single instance will be trained and accessed only on a single center point, which reduces the calculation. 
However, it also restricts the perception field of regression and affects reliability.
To overcome these disadvantages, we apply the approach used by Eskil et al.~\cite{jorgensen2019monocular} and Krishna et al.~\cite{krishnan2016vehicle}. 
Instead of relying on a single point, a \textbf{R}eference \textbf{A}rea (\textbf{RA}) based on the 2D center point is defined within the 2D bounding box, whose width and height are set accordingly with a proportional value $\gamma$.
All values within this area contribute to regression and classification.
If RAs overlap, the area closest to the camera dominates, since only the closest instance is completely visible on the monocular image.
An example using RAs is shown in Figure~\ref{fig:ra}.
During inference all predictions in the related RA will be weighted equally.

\section{Experiments}
\subsection{Implementation Details}
We performed experiments on the KITTI object detection benchmark~\cite{geiger2012we}, which contains 7481 training images with annotation and calibration.
KITTI was chosen from the public available datasets, e.g.~\cite{Cordts2016Cityscapes},\cite{geiger2012we},\cite{aev2019},\cite{apolloscape_arXiv_2018}, since it is the widely used benchmark by previous works for monocular 3D detection, including our baseline, CenterNet.  
All instances are divided into easy, moderate and hard targets according to visibility in the image~\cite{geiger2012we}.
We follow the standard training/validation split strategy in literature~\cite{chen20153d}, which leads to 3712 images for training and 3769 images for validation.
Like most previous work, and in particular CenterNet, we only consider the ``Car'' category.
By default, KITTI evaluates 2D and 3D performance with AP at eleven recalls from $0.0$ to $1.0$ at different IoU thresholds.
For a fair comparison with CenterNet, parameters stay unchanged if not stated otherwise. 
In particular we keep the modified Deep Layer Aggregation (DLA)-34~\cite{yu2018deep} as the backbone. 
Regarding different approaches, we add specific head layers, which consist of one $3\times 3$ convolutional layer with 256 channels, ReLu activation and a $1\times 1$ convolution with desired output channels at the end.
We trained the network from scratch in PyTorch~\cite{paszke2017automatic} on 2 GPUs (1080Ti) with batch sizes 7 and 9. 
We trained the network for 70 epochs with an initial learning rate of $1.25e^{-4}$, which drops by a factor of 10 at 45 and 60 epochs if not specified otherwise.

\subsection{Offset3D}
With Offset3D we bridge the gap between 2D and 3D center points by adding 2 specific layers to regress the offset $\mathbf{\Delta c}^i$. 
For demonstration we perform an experiment, which is indicated as {\em CenterNet(ct3d)} in Table~\ref{tab:sota}.
It models the object with a projected 3D center point with 4 distances to boundaries. 
The visible object, whose 3D center point is out of the image, is ignored during training.
As the second row in Table~\ref{tab:sota} shows, for easy targets {\em CenterNet(ct3d)} increases the BEV AP by $48.6\%$ and the 3D AP by $104.6\%$ compared to the baseline of CenterNet.
This is achieved by the proper decoding of 3D parameters based on an appropriate 3D center point.
However, as discussed in Section~\ref{sec:offset3d}, simply modeling an object with a 3D center will hurt 2D performance, since some 3D centers are not attainable, although the object is still partly visible in the image.

The Offset3D approach is able to balance the trade-off between a tightly fitting 2D bounding box and a proper 3D location.
Center3D is showing that the regression of offsets is regularizing the spatial center, while also preserving the stable precision in 2D (the 7th row in Table~\ref{tab:sota}).
With a higher learning rate of $2.4e^{-4}$, BEV AP for easy targets improves from $31.5\%$ to $50.7\%$, and 3D AP increases from $19.5\%$ to $43.1\%$, which performs comparably to the state of the art.
Since Offset3D is also the basis for all further experiments, we treat the performance as our new baseline for comparison.

\subsection{LID}
\label{sec:exp_lid}
We first implement and adjust the DORN~\cite{fu2018deep} approach for depth estimation instance-wise rather than pixel-wise. 
Following DORN we add a shift $\xi$ to both distance extremum $d_\text{min}^*$ and $d_\text{max}^*$ to ensure $d_\text{min} = d_\text{min}^* + \xi = 1.0$.
In addition, we perform experiments for our LID approach to demonstrate its effectiveness.
We set the number of bins to 80, and add a single head layer to regress the residuals of the discretization.
Hence, for depth estimation, we add head layers to generate the output features $\hat D \in \mathbb{R}^{\frac{W}{R} \times \frac{H}{R} \times 161}$, while Offset3D generates an output feature $\hat D \in \mathbb{R}^{\frac{W}{R} \times \frac{H}{R} \times 1}$.
For both approaches the depth loss weight $\lambda_\text{dep}$ yields the best performance.
We compare the results with our new baseline Offset3D with the same learning rate of $1.25e^{-4}$. 

Table~\ref{tab:lid} shows both LID and SID with different number of bins improved even instance-wise with additional layers for ordinal classification, when a proper number of bins is used. 
Our discretization strategy LID shows a considerably higher precision in 3D evaluation, comprehensively when 80 and 100 bins are used.
A visualization of inferences of both approaches from BEV is shown in Figure~\ref{fig:lid}.
LID only preforms worse than SID in the 40 bin case, where the number of intervals is not enough for instance-wise depth estimation.
Furthermore we verify the necessity of the regression of residuals by comparing the last two rows in Table~\ref{tab:lid}.
The performance of LID in 3D will be deteriorate drastically if this refinement module is removed..

\begin{table}[!t]
	\centering
	\tabcolsep=0.1cm
	\begin{tabular}{l c c c c c c c}
		\hline
		& & \multicolumn{3}{c}{BEV AP} & \multicolumn{3}{c}{3D AP} \\
		& Bin & Easy & Mode & Hard & Easy & Mode & Hard \\
		\hline
		Offset3D & - & 47.6 & 37.6 & 32.4 & 38.0 & 30.8 & 29.4 \\
		\hline
		\hline
		SID & 40 & 33.4 & 27.6 & 26.9 & 26.7 & 24.4 & 21.1 \\
		LID & 40 & 31.5 & 25.6 & 24.9 & 24.7 & 22.7 & 19.4 \\
		\hline
		\hline
		SID & 100 & 47.6 & 37.5 & 32.3 & 39.6 & 33.8 & 29.4 \\
		LID & 100 & \underline{50.2} & \underline{39.2} & \textbf{33.9} & \underline{41.5} & \textbf{35.6} & \textbf{31.3} \\
		\hline
		\hline
		SID & 80 & 48.7 & 37.9 & \underline{32.9} & 40.4 & 34.3 & 30.0 \\
		LID & 80 & \textbf{51.3} & \textbf{39.3} & \textbf{33.9} & \textbf{44.0} & \underline{35.0} & \underline{30.6} \\
		LID / -res & 80 & 37.1 & 33.0 & 29.2 & 31.9 & 26.7 & 25.8 \\
		\hline
	\end{tabular}
	\caption{Experimental results of LID. We show the comparison between SID and LID, the influence of different bins and runtime (RT). {\em -res} indicates no regression of residuals as an ablation study. APs are given in percentage.}
	\label{tab:lid}
\end{table}

\subsection{DepJoint and Reference Area}
In this section, we evaluate the performance of DepJoint and RA, and finally discuss the compatibility of both approaches.

\subsubsection{Reference Area}
\label{sec:ra}
The RA enriches the cues for inference and utilizes more visible surface features of instances.
Focusing especially on 3D performance, we evaluate RA over depth estimation and explore the sensitivity to the size of the RA.
Table~\ref{tab:RA} shows the improvement of models supported by the RA for 3D detection. 
As we can see in Table~\ref{tab:RA}, performance does not simply improve with higher proportion scale, $\gamma$, since the best AP is achieved with $\gamma=0.4$ .
We attribute this characteristic to the misalignment of estimated 2D bounding boxes and the position of the RA.
The output features of depth in the RA respond exactly to the input pixels on their location.
However, the localization of the RA during inference is not the same as in ground truth.
As a result, a shift of the estimated location or the size of the RA involves the feature values, which in fact belong to another instance, especially when objects overlap.
A weighted average during decoding could reduce but generally not eliminate this error.
Furthermore, we also determine the effectiveness of RA over the offset prediction $\mathbf{\Delta C}$. 
Therefore, the choice of loss weight to balance the decreasing losses between tasks is tricky.
Finally $\lambda_\text{off} = 0.025$ performs best.
A smaller value of $\lambda_\text{off}$ downweights the magnitude of the offset loss and does not influence the training of other losses.
Instead, it will be fine tuned in late period.

\begin{table}[t]
	\centering
	\tabcolsep=0.09cm
	\begin{tabular}{l c c c c c c c}
		\hline
		& & \multicolumn{3}{c}{BEV AP} & \multicolumn{3}{c}{3D AP} \\
		\hline
		RA & $\gamma$ & Easy & Mode & Hard & Easy & Mode & Hard \\
		\hline
		\hline
		dep & 0.2 & 50.8 & 40.2 & 35.4 & 46.1 & 37.1 & 32.9 \\
		dep & 0.4 & \underline{52.8} & \underline{41.5} & 35.5 & \underline{48.4} & \textbf{38.7} & \underline{33.2} \\
		dep & 0.6 & 52.7 & 41.3 & 35.5 & 47.5 & 37.5 & 32.9 \\
		dep & 0.8 & 50.7 & 39.2 & 33.7 & 45.6 & 35.3 & 30.9 \\
		dep & 1.0 & 50.9 & 39.2 & 32.4 & 46.3 & 32.4 & 30.7 \\
		\hline
		\hline
		dep+off(1) & 0.4 & 48.2 & 37.1 & 32.0 & 41.9 & 30.9 & 29.2 \\
		dep+off(0.1) & 0.4 & 52.6 & 41.1 & \underline{39.6} & 47.0 & 37.5 & 32.9 \\
		dep+off(0.025) & 0.4 & \textbf{54.3} & \textbf{41.9} & \textbf{40.3} & \textbf{48.5} & \underline{38.0} & \textbf{33.7} \\
		\hline
	\end{tabular}
	\caption{Experimental results of Reference Area (RA). The first column indicates which sub-tasks are supported by RA. The default weighting of {\em dep} or {\em off} loss during training are set to 1, if not specified otherwise in $(\cdot)$. The parameter $\gamma$ ranged from 0.2 to 1.0 and defines the scale value of RA regarding to a 2D bounding box. APs are given in percentage.}
	\label{tab:RA}
\end{table}

\begin{table}[t]
	\centering
	\tabcolsep=0.09cm
	\begin{tabular}{c c c c c c c c}
		\hline
		& & \multicolumn{3}{c}{BEV AP} & \multicolumn{3}{c}{3D AP} \\
		\hline
		$\lambda_\text{dep}$ & $\alpha/\beta$ & Easy & Mode & Hard & Easy & Mode & Hard \\
		\hline
		1 & 0.7/0.3 & 52.3 & \textbf{41.5} & \textbf{35.3} & \underline{46.4} & \underline{37.3} & \underline{32.5} \\
		0.1 & - & \underline{52.9} & 40.6 & 35.0 & 44.5 & \textbf{37.5} & \textbf{32.7} \\
		0.05 & - & 51.7 & 39.7 & 34.6 & 43.1 & 32.6 & 31.3 \\
		\hline
		\hline
		0.1 & 0.6/0.4 & 49.9 & 39.6 & 34.1 & 44.7 & 35.5 & 30.8 \\
		- & 0.8/0.2 & 47.9 & 38.7 & 33.6 & 40.3 & 31.9 & 30.9 \\
		- & 0.9/0.1 & 48.4 & 37.8 & 32.7 & 41.4 & 31.2 & 29.4 \\
		\hline
		\hline
		- & 0.2/0.2 & 48.6 & 37.5 & 33.3 & 42.1 & 32.9 & 30.9 \\
		- & 0.3/0.3 & \textbf{53.2} & \underline{41.2} & \underline{35.1} & \textbf{47.4} & 36.4 & 32.3 \\
		- & 0.4/0.4 & 51.5 & 40.3 & 34.1 & 45.5 & 36.0 & 31.2 \\
		- & 0.5/0.5 & 42.9 & 35.9 & 31.9 & 37.7 & 29.3 & 28.3 \\
		\hline
	\end{tabular}
	\caption{Experimental results of DepJoint with the support of RA. $\lambda_\text{dep}$ indicates the weighting of depth loss during training. The dependence on $\alpha/\beta$ is shown, which represents the threshold scale of first/second bins regarding to $d_\text{max}$. Where parameters have the same values as the previous row it is marked as - in Table. APs are given in percentage.}
	\label{tab:Joint}
\end{table}

\subsubsection{DepJoint}
In the following sections we combine the DepJoint approach with the concept of RA.
As discussed above we set $\gamma=0.4$ as default for RA.
Additionally, we apply $d_\text{min} = 0$ and $d_\text{max} = 60$ for all experiments.
Table~\ref{tab:Joint} shows that DepJoint is more robust against loss weighting during training in comparison with Eigen's approach. 
As introduced in Section~\ref{sec:depjoint}, we can divide the whole depth range into two overlapping or associated bins.
For the associated strategy, the thresholds $\alpha/\beta$ of $0.3/0.3$ and $0.4/0.4$ show the best performance for the following reason:
usually more distant objects show less visible features in the image.
Hence, we want to set both thresholds a little lower, to assign more instances to the distant bins and thereby suppress the imbalance of visual clues between the two bins.

\subsection{Comparison to the State of the Art}
\label{sec:sota}
Table~\ref{tab:sota} shows the comparison with state-of-the-art monocular 3D detectors on the KITTI dataset. 
The LID approach follows the settings described in Section~\ref{sec:exp_lid}. 
Different from the experiments above, all RAs are applied for both depth $d$ and offset $\Delta C$ estimation ($\gamma= 0.4$). 
Our first model, {\em Center3D(+ra)}, combines the reference area approach with Eigen's transformation for depth estimation.
The loss weights are set to $\lambda_\text{dep} = 1$ and $\lambda_\text{offset} = 0.025$, while the learning rate is $2.4e^{-4}$.
Finally, we evaluated the performance of a model employing both DepJoint and RA approaches, {\em Center3D(+dj+ra)}. 
The training with $\lambda_\text{dep} = 1$, $\lambda_\text{offset} = 0.1$ and learning rate $1.25e^{-4}$ yields the best experimental result.

As Table~\ref{tab:sota} shows, all Center3D models perform at comparable 3D performance with respect to the best approaches currently available.
{\em Center3D(+dj+ra)} achieved state-of-the-art performance with BEV APs of $55.8\%$ and $42.8\%$ for easy and moderate targets, respectively. 
For hard objects in particular, {\em Center3D(+ra)} outperforms all other approaches.
The RA improved the BEV AP by $4.9\%$ (from $35.3\%$ to $40.2\%$) and $1.0\%$ (from $33.0\%$ to $34.0\%$).

Center3D preserves the advantages of an anchor-free approach.
It performs better than most other approaches in 2D AP, especially on easy objects.  
Most importantly, it infers on the monocular input image with the highest speed (around three times faster than M3D-RPN, which performs similarly to Center3D in 3D).
Therefore, Center3D is able to fulfill the requirement of a real-time detection.
Additionally, it achieved the best trade-off between speed and performances not only in 2D but also in 3D.

\section{Conclusion}
In this paper we introduced Center3D, a one-stage anchor-free monocular 3D object detector, which models and detects objects with center points of 2D bounding boxes. 
We recognize and highlight the importance of the difference between centers in 2D and 3D by regressing the offset directly, which transforms 2D centers to 3D centers. 
In order to improve depth estimation, we further explored the effectiveness of joint classification and regression when only monocular images are given.
Both classification-dominated (LID) and regression-dominated (DepJoint) approaches enhance the AP in BEV and 3D space.
Finally, we employed the concept of RAs by regressing in predefined areas, to overcome the sparsity of the feature map in anchor-free approaches.
Center3D performs comparably to state-of-the-art monocular 3D approaches with significantly improved runtime during inference.
Center3D achieved the best trade-off between 3D precision and inference speed.

\section{Acknowledgement}
We gratefully acknowledge Heinz Koeppl for his support of this work as well as Christian Wildner for his advice. Furthermore, we thank Andrew Chung and Mentar Mahmudi for helpful comments and discussions.

\clearpage
\begin{spacing}{1.1}
{
	\bibliographystyle{ieee}
	\bibliography{ref}
}
\end{spacing}

\end{document}